\pdfoutput=1

\documentclass[11pt]{article}

\usepackage{EMNLP2022}

\usepackage{times}
\usepackage{latexsym}
\usepackage{hyperref}

\usepackage[T1]{fontenc}

\usepackage[utf8]{inputenc}

\usepackage{microtype}

\usepackage{enumitem}
\usepackage{multirow}

\usepackage{graphicx}
\graphicspath{{images/}}

\title{Garden Path Traversal in GPT-2}

\author{
  William Jurayj \\
  Brown University \\
  $\texttt{william@jurayj.com}$ \\\And
  William Rudman \\
  Brown University \\
  $\texttt{william\_rudman@brown.edu}$ 
  \\\AND
  Carsten Eickhoff \\
  Brown University \\
  $\texttt{c.eickhoff@acm.org}$
}

\begin{document}
\maketitle
\begin{abstract}
In recent years, large-scale transformer decoders such as the GPT-x family of models have become increasingly popular. Studies examining the behavior of these models tend to focus only on the output of the language modeling head and avoid analysis of the internal states of the transformer decoder. In this study, we present a collection of methods to analyze the hidden states of GPT-2 and use the model's navigation of garden path sentences as a case study. To enable this, we compile the largest currently available dataset of garden path sentences. We show that Manhattan distances and cosine similarities provide more reliable insights compared to established surprisal methods that analyze next-token probabilities computed by a language modeling head. Using these methods, we find that negating tokens have minimal impacts on the model's representations for unambiguous forms of sentences with ambiguity solely over what the object of a verb is, but have a more substantial impact of representations for unambiguous sentences whose ambiguity would stem from the voice of a verb. Further, we find that analyzing the decoder model's hidden states reveals periods of ambiguity that might conclude in a garden path effect but happen not to, whereas surprisal analyses routinely miss this detail.
\end{abstract}

\begin{figure}[!ht]
    \centering
    \includegraphics[width=7.5cm]{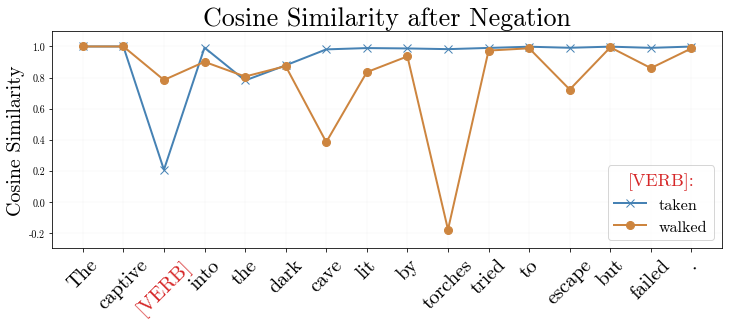}
    \includegraphics[width=7.5cm]{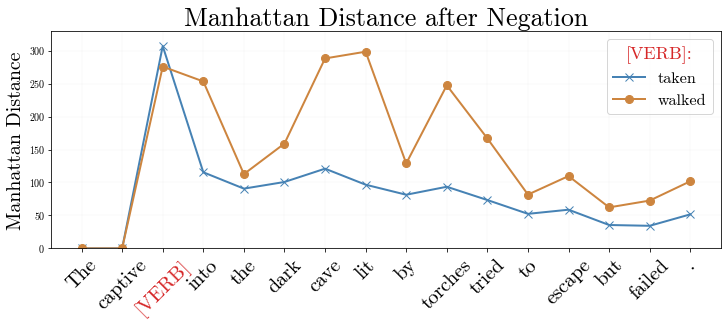}
    \includegraphics[width=7.5cm]{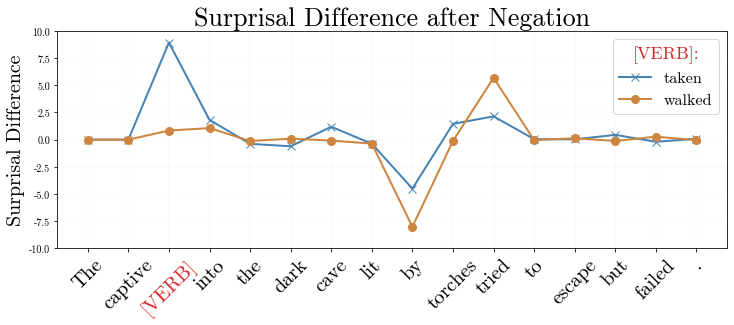}
    \caption{Hidden state relations (Top: cosine similarity, Middle: Manhattan distance, Bottom: surprisal difference) between negated and non-negated forms of garden path and unambiguous sentences. The ambiguous verb ``walked'' primes the effect later in the sentence, while the unambiguous ``taken'' avoids it. The verb ``lit'' introduces a similar ambiguity, which hidden state metrics show but surprisal misses because the next word ``by'' does not trigger a garden path effect.}
    \label{fig:cos_test}
\end{figure}

\begin{figure*}[h!]
    \centering
    \includegraphics[width=15cm]{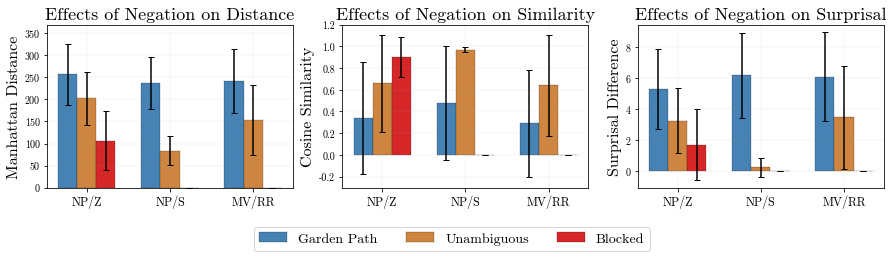}
    \caption{Left: Average Manhattan distances between sentence types and their negated forms. Center: Average Cosine similarities between sentence types and their negated forms. Right: Average surprisal differences between sentence types and their negated forms. Manhattan distances exhibit less variability than either cosine similarities or surprisal differences}
    \label{fig:results}
\end{figure*}

\section{Introduction}

OpenAI's release of GPT-3 marked a major step in the field of massive language models, whose ability to generate news articles indistinguishable from those written by humans provides a salient example of the many social and political implications of these models \cite{GPT3, universal-adversarial-triggers, gpt2-flat-earth}. Within 2 years of BERT's release, over 150 studies have investigated BERT's structure, exploring how its internal representations enable powerful and flexible language comprehension \cite{BERT-geometry, BERT-dark-secrets, BERT-classical-nlp, BERTology-primer}.  A few such studies include a decoder model in the set of models examined, but do not specifically design their analyses around this type of architecture \cite{what-do-you-learn-from-context, linguistic-knowledge-contextual-representation}. Meanwhile, studies exploring GPT models alone tend to focus on properties of text generated from its language modeling head, and do not analyze the internal representations of the model in depth \cite{gpt2-flat-earth, GPT3}.

The few studies that explore the hidden states of GPT-2 suggest an under-utilization of its massive latent space as representations are dominated by the presence of rogue dimensions \cite{gpt-cone, gpt-isotropy, isoscore, gpt-rogue-dims}. As massive decoder models become more ubiquitous and powerful, it will become ever more important to understand the internal processes by which they generate content so they can be streamlined and improved upon.

In this paper, we use garden path traversal as a case study to demonstrate the value of directly analyzing properties of the embedding space in transformer decoder models. A garden path sentence is one where the parse that a reader expects at some point within the sentence is proven incorrect by the end of the sentence. We choose to explore this syntactic effect specifically because we believe the intuitive reaction a human has when reading such sentences provides a helpful frame for analyzing the behavior of a neural language model experiencing the same effect.

We expect that by looking at the hidden states from which next word likelihoods are computed, we can observe the same patterns that surprisal analysis reveals, while uncovering more nuanced trends that surprisal misses because it depends on the joint distribution of the hidden state and the next word. Moreover, we expect that Manhattan distances will exhibit less variance in the effect of negating a given sentence type than either surprisals or cosine similarities after a zero-mean transformation, because Manhattan distances are resilient to extreme values in a single dimension \cite{euclidean-manhattan}. On the other hand, the next word likelihoods used to compute surprisal tend to depend heavily on these dimensions, while the zero-mean translation required to create meaningful angular differences around the origin mean that a few extreme dimensions expose cosine similarities to similarly high variances \cite{gpt-rogue-dims}.

By analyzing how GPT-2 sequentially embeds tokens in space, we are able to identify how GPT-2 internally handles different garden path effects. Specifically, we show that GPT-2 recognizes potential but unrealized garden path effects using metrics that examine the model's hidden states, whereas surprisal analysis fails to reveal this finding. We argue that analysis of a decoder model's hidden states enables more robust analysis than can be done using the next word likelihoods alone, which themselves are distilled from these hidden states. The contributions of this study are as follows:

\setlist{nolistsep}
\begin{itemize}[noitemsep]
    \item to introduce the largest and most diverse dataset of garden path sentences currently available, along with construction functions to negate or extend the effect within each sentence,
    \item to demonstrate the advantage of analyzing syntactic properties such as garden path effects by examining geometric relationships between vectors in GPT-2's hidden states using Manhattan distance and cosine similarity,
    \item to motivate further study of the hidden states of transformer decoders as a more thorough alternative to the surprisal-based methods that are typically used to analyze language models.\footnote{Code available at \href{https://github.com/wjurayj/garden-path-gpt2}{https://github.com/wjurayj/garden-path-gpt2}}
\end{itemize}

\begin{table*}[t]
    \begin{tabular}{c | c | c} 
     \hline
     \hline
     \textbf{Sentence Type} & \textbf{Sentence Form} & \textbf{Sentence} \\ [0.5ex] 
     \hline
     \hline
        \multirow{5}{4em}{\textbf{NP/Z}} & Garden Path & When the dog scratched the vet {\color{red}\underline{took off}} the muzzle.\\
        & Negated & When the dog scratched, the vet {\color{red}\underline{took off}} the muzzle.\\
        & Blocked & When the dog scratched his owner the vet
        {\color{red}\underline{took off}} the muzzle.\\
        & Unambiguous & When the dog struggled the vet
        {\color{red}\underline{took off}} the muzzle.\\

    \hline
        \multirow{4}{4em}{\textbf{NP/S}} & Garden Path & The coach discovered the player {\color{red}\underline{tried}} to show off all the time.\\
        & Negated & The coach discovered that the player {\color{red} \underline{tried}} to show off all the time.\\
        & Unambiguous & The coach thought the player {\color{red}\underline{tried}} to show off all the time.\\
    \hline
        \multirow{4}{4em}{\textbf{MV/RR}} & Garden Path & The horses raced past the barn {\color{red}\underline{fell}} into a ditch.\\
        & Negated & The horses that were raced past the barn {\color{red}\underline{fell}} into a ditch.\\
        & Unambiguous & The horses ridden past the barn {\color{red}\underline{fell}} into a ditch.\\
    \hline
    \hline
    \end{tabular}
    \caption{\label{tab:examples} Forms of NP/Z, NP/S, and MV/RR sentences included in our dataset, with the verb that triggers or would trigger the garden path effect underlined in red. Note that all of the perturbations can be combined to avoid the garden path effect, except for the blocked and unambiguous forms of the NP/Z sentence.}
\end{table*}

\subsection{Related Work}

Many studies into GPT or BERT involve fine-grained analyses of how the model handles specific syntactic phenomena, such as the garden path effect. Consider the sentence:
\begin{center}
    ``Even though the girl phoned[,] the instructor was very upset with her for missing a lesson.''
\end{center}
Without the comma, most readers will assume ``the instructor'' is the direct object of the verb ``phoned'', rather than the subject of the main clause's verb phrase, ``was very upset'' \cite{Schijndel2019NeuralNS}. Adding the comma immediately disqualifies the incorrect parse, nullifying the garden path effect. This method of preventing the effect is referred to as ``negation'' throughout this paper.

Analysis of garden path traversal is typically done by comparing the surprisal, or negative log likelihood, of the token that would trigger the garden path effect between garden path and negated sentences. The surprisal that a token induces from a language model can intuitively be understood to measure the amount of information that token adds to that model's representation of the sentence, as measured by the inverse of the degree to which the model anticipated that token. This is calculated using a language modeling head on top of GPT-2, and does not directly analyze the internal representations of the model from which these likelihoods are computed.

Previous studies into the navigation of these sentences find that sufficiently large models' relative surprisals at the disambiguating token between garden path and negated sentences show recognition of the garden path effect. However, these models systematically underestimate the magnitude of the effect observed in humans, suggesting that human recovery from an incorrect parse involves more than just the triggering token's lack of predictability \cite{vanSchijndel2021SingleStagePM, vanSchijndel2018ModelingGP}. Further, using surprisal comparisons, \citet{hu-etal-2020-systematic} show that GPT-2 recognizes garden path effects less successfully or consistently than smaller recurrent language models. 

OpenAI has not released GPT-3's source code and parameters, so we instead analyze its predecessor GPT-2, which uses an almost identical architecture at a much smaller scale (1.5b parameters). Nonetheless, the methods we use to explore GPT-2's traversal of garden path effects can be easily generalized to study any decoder-based model.

\section{Methods}
\subsection{Garden path sentence generation}
The dataset used for these experiments builds on the combination of the NP/Z and NP/S sentences from \citet{Grodner2003} and the NP/Z and MV/RR sentences from \citet{psycholinguistic-subjects}, originally taken from \citet{Staub2007-rn} and \citet{Tabor2004-ev}, and consists of 43 NP/Z sentences, 20 NP/S sentence, and 20 MV/RR sentences. Instead of building out side-by-side datasets of each type of sentence, however, we store the components of these sentences in tabular files, and include scripts to construct these sentences in various forms similar to those used by \citet{psycholinguistic-subjects}. Each sentence has a \textit{garden path} and an \textit{unambiguous} form, depending on whether the first verb allows for an ambiguous parse. Each of these forms can be \textit{negated} with the addition of one or two tokens, which nullifies the garden path effect in an ambiguous sentence but makes no semantic difference in an unambiguous sentence. We provide this as a template to be extended indefinitely to meet the needs of future research. Examples of each sentence type's possible forms can be found along with a detailed description of these effects in Table~\ref{tab:examples}.

\begin{figure}[!ht]
    \centering
    \includegraphics[width=7.5cm]{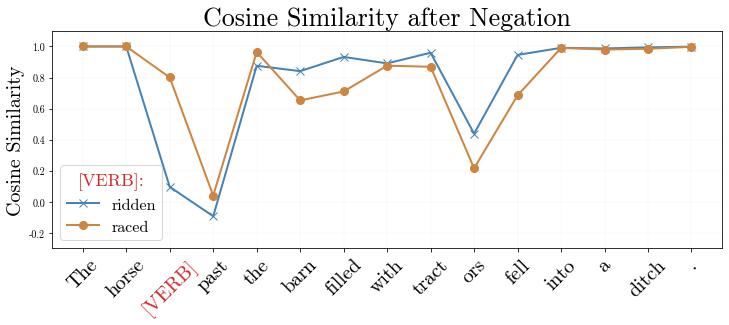}
    \includegraphics[width=7.5cm]{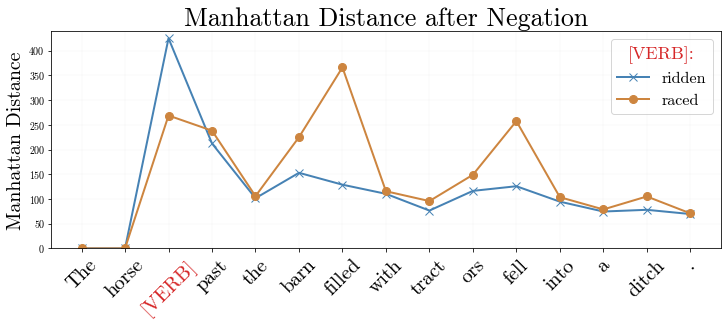}
    \includegraphics[width=7.5cm]{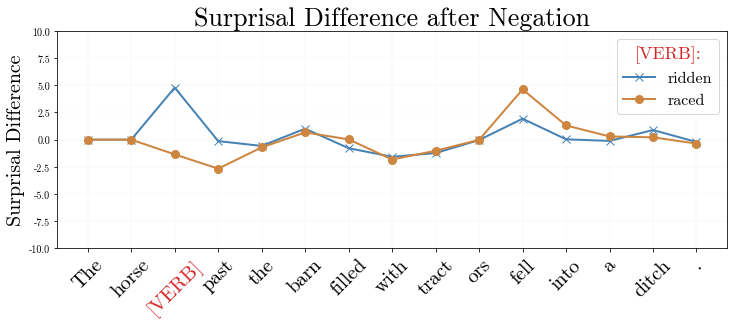}
    \caption{Hidden state relations (Top: cosine similarity, Middle: Manhattan distance, Bottom: surprisal difference) between negated and non-negated forms of garden path and unambiguous sentences. The ambiguous verb ``raced'' primes the effect later in the sentence, while the unambiguous ``ridden'' avoids it. Like in Figure~\ref{fig:cos_test}, all metrics catch the garden path effect at the verb ``fell'', but only cosine similarity and Manhattan distance anticipate the possible effect at ``filled''}
    \label{fig:trajectory1}
\end{figure}

\subsubsection{NP/Z sentences}
NP/Z is short for Noun Phrase/Zero complement. These are sentences where the first verb appears to take on a Noun Phrase as its direct object, but subsequently is revealed to have no (Zero) direct object at all \cite{psycholinguistic-subjects}. The garden path effect in these sentences is caused by ambiguity about whether the verb of the leading subordinate clause has a direct object. These sentences have an additional \textit{blocked} form, which nullifies its garden path effect by adding an explicit direct object to the leading verb. This is considered one of the stronger types of garden path effects, with an average increase in human reading time of 152 ms \cite{reading-times}.

The first NP/Z sentence in Table~\ref{tab:examples} evokes a garden path effect because the reader initially expects that ``the vet'' is the direct object of ``scratched''; The negated form avoids the effect by using a comma to indicate the separation between the two clauses. The blocked form avoids the effect by adding the direct object ``his owner'' to block the ambiguity that triggers the effect, while the unambiguous form avoids the effect by replacing the transitive verb ``scratched'' with the intransitive verb ``struggled'' to avoid ambiguity around the verb's direct object.

Our dataset includes 43 distinct NP/Z sentences, and includes scripts allowing a user to easily transform these into unambiguous or blocked sentences. Moreover, each sentence has the option to include a negation, and an extension so as to increase the duration of the ambiguity.

\begin{figure*}[h!]
    \centering
    \includegraphics[width=15cm]{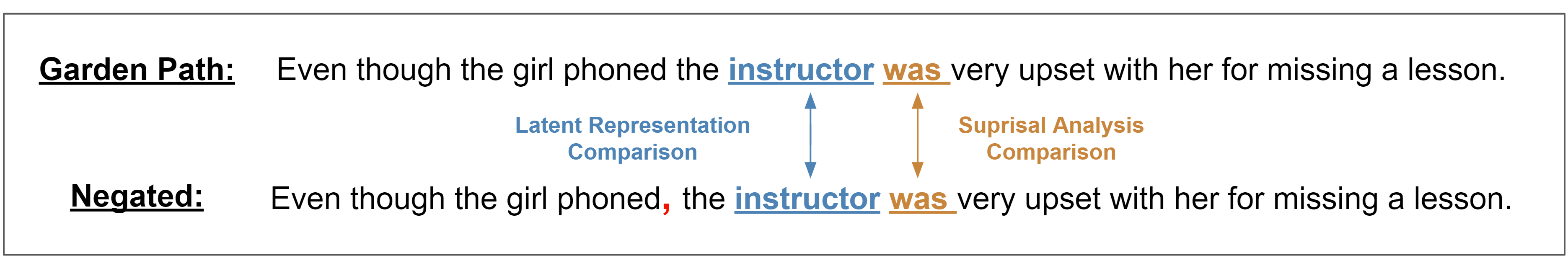}
    \caption{Method for comparing latent space metrics (cosine similarity, Manhattan distance) against surprisal difference.}
    \label{fig:method}
\end{figure*}

\subsubsection{NP/S sentences}

NP/S is short for Noun Phrase/Sentential complement. These are sentences where the first verb appears to take the Noun Phrase as its direct object, but subsequently is revealed to have a sentence-like object as its complement \cite{vanSchijndel2018ModelingGP}. The garden path effect in these sentences is caused by ambiguity about whether the noun following the main clause's verb is that verb's direct object. This is considered one of the weaker types of garden path effects, with an average increase in human reading times of 50 ms \cite{reading-times}.

The first NP/S sentence in Table~\ref{tab:examples} evokes a garden path effect because the reader expects that ``the player'' is the direct object of the verb 'discovered' until the word ``tried'' reveals that it is her propensity to show off that the coach is discovering. The negated form avoids the effect by adding ``that'' before ``the player'' to eliminate the possibility that 'the player' is the verb's direct object. The unambiguous form avoids the effect altogether by using the verb ``thought'', which could not allow a person to be its direct object. Our dataset includes 20 distinct NP/S sentences, each of which can be negated,  unambiguous, extended, or any combination thereof.

\subsubsection{MV/RR sentences}
MV/RR is short for Main Verb/Reduced Relative. These are sentences where prior to a disambiguator, the ambiguous verb could either be the main verb of the sentence or a verb that introduces a reduced relative clause \cite{psycholinguistic-subjects}. The garden path effect in these sentences is caused by ambiguity about whether the past-tense verb of the leading subordinate clause is a past participle or the main verb of the sentence. This effect is considered stronger than that of an NP/S sentence, but reading time data to compare it with the other sentence types is not available.

The first MV/RR sentence in Table~\ref{tab:examples} evokes the garden path effect because the reader assumes ``raced'' is the main verb of the sentence, while the negated form negates this ambiguity by clarifying that ``raced past the barn'' is a descriptor for the horses rather than the main clause itself. Note that in some examples, the negating tokens are ``who were'' instead of ``that were'', but in both cases these tokens serve to un-reduce the relative participle. The unambiguous form avoids ambiguity altogether by replacing the ambiguous ``raced'' with the unambiguously passive ``ridden''. Our dataset includes 20 distinct MV/RR sentences, each of which can be negated, rendered unambiguous, extended, or any combination thereof.

\subsection{Experimental design}
The general structure of the tests we run is inspired by \citet{psycholinguistic-subjects} and \citet{hu-etal-2020-systematic}. The key difference is that, where previous studies compare the model's surprisal at the disambiguating word, we examine the model's hidden state prior to this word. Figure~\ref{fig:method} shows a visualization of this approach.

We compare each sentence to its negated form, computing the vector differences and cosine similarities between each token and its counterpart in the negated form (omitting the token[s] that were added to negate the garden path effect in that sentence type from the pairing process) after re-centering embeddings around the origin. We use Manhattan distance over Euclidean distance to compute scalars from the vector differences between sentences as is generally preferred in high dimensional spaces, where Euclidean distances are sensitive to the dimensions with the largest values \cite{euclidean-manhattan}. Cosine similarities are computed after re-centering all vectors so that the distribution has a mean of zero, which prevents the metric from defaulting to near-maximum values and allows it to measure the true directional changes between vectors \cite{isoscore}. These side-by-side metrics are generated for all sentences' garden path and unambiguous forms, as well as for the blocked form of the NP/Z sentences.

We expect to see larger distances and lower similarities upon negation in garden path sentences than in unambiguous or blocked sentences. In the garden path sentences the negating tokens help to resolve some ambiguity, whereas in an already unambiguous sentence they will contribute minimally to the sentence's meaning prior to the triggering token.

\section{Results \& Discussion}

Our analysis reveals several properties of GPT-2's experience of the garden path effect. Across all sentence types, Manhattan distances and cosine similarities show that the model reacts more heavily to negation of garden path sentences than it does to these sentences' unambiguous counterparts, as is reflected by surprisal analyses done here and in previous studies \cite{sarti-thesis}.

Although our surprisal baselines mirror the trends seen in Manhattan distance, using Manhattan distances provides more consistent results compared to surprisal analysis. Our results demonstrating exceedingly high variance in the surprisal analysis is in line with the findings of \citet{hu-etal-2020-systematic}, who use surprisal to show that GPT-2 performs especially poorly and inconsistently on garden path effects. On the other hand, the high-level trends we expected to see are present across all metrics, with negation causing a less pronounced difference in unambiguous and blocked sentences than it does in garden path sentences. Whereas Figure~\ref{fig:results} shows Manhattan distances to have relatively low variance compared to the other metrics we examine, cosine similarity and surprisal suffer from very high variances within each sentence form. We believe that this is due to Manhattan distance's resistance to GPT-2's rogue dimensions, which dominate cosine similarities after the zero-mean transformation because even relatively minor differences in a few of these dimensions will have a much more substantial effect on the angular difference than more major differences in many smaller dimensions would \cite{gpt-rogue-dims, euclidean-manhattan}.

Figure~\ref{fig:cos_test} illustrates how these high level trends appear within a given sentence. Here, the verb ``tried'' triggers the garden path effect, which is directly preceded by a spike in Manhattan distance and a dip in cosine similarity between the garden path and negated sentence forms as the model anticipates different continuations: in the garden path sentence the model is likely to predict some sort of punctuation or conjunction to end the clause, while in the negated form the model expects a verb to complete the clause that the ambiguous verb is subordinate to.

An interesting property of the specific example in Figure~\ref{fig:cos_test}, \textcolor{red}{``The captive walked into the dark cave lit by torches tried to escape but failed.''} is that the verb ``lit'' also triggers a momentary garden path effect; the sentence could, for instance, simply continue, \textcolor{red}{``The captive walked into the dark cave lit the torches.''} In the first case, the verb ``lit'' is a reduced relative of ``that was lit'', which refers to the cave, whereas in the second case the captive is the subject of ``lit'', which is the main verb of the sentence and thus would trigger a garden path effect. Of course, the immediate next word ``by'' eliminates this possibility, which a human reader notices quickly enough that they do not experience the effect, but the decoder's causal attention mask allows it no such foresight.

This possibility is thus worth considering because it helps explain why there is a spike in Manhattan distance and a dip in cosine similarity at the preceding word, ``cave''.
We believe the relative shallowness of the dip in cosine similarity before ``lit'' is due to the possible MV/RR ambiguity of the word, since even in the garden path case where punctuation can be expected, a verb such as ``lit'' can preempt an adjectival clause as it does here (``lit by torches''). This puts it in a curious superposition between introducing and triggering a garden path effect, at least until the next word ``by'' resolves this ambiguity.
Notably, in the garden path form neither hidden state metric returns to its baseline value until after the verb ``lit'', because the model expects this verb leads a subordinate clause whereas in the negated form it considers both possibilities.

\begin{figure}[!ht]
    \centering
    \includegraphics[width=7.5cm]{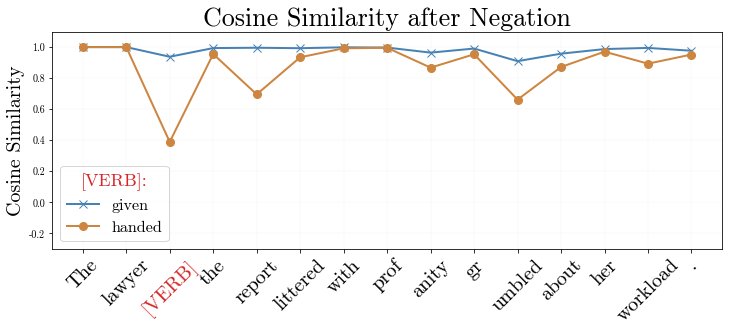}
    \includegraphics[width=7.5cm]{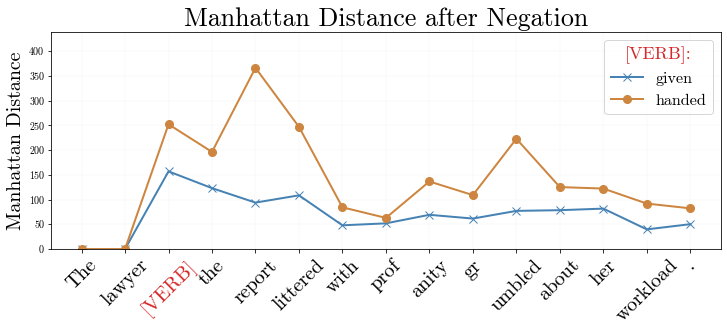}
    \includegraphics[width=7.5cm]{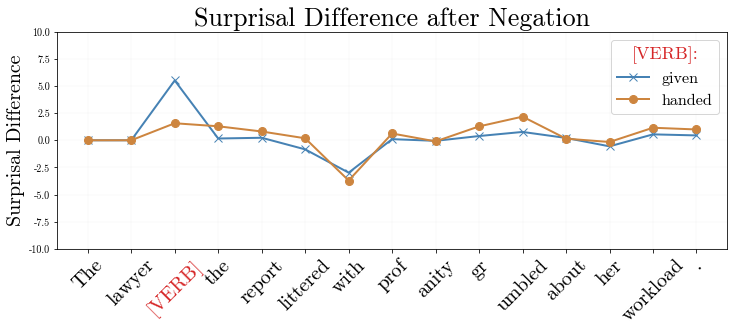}
    \caption{Hidden state relations (Top: cosine similarity, Middle: Manhattan distance, Bottom: surprisal difference) between negated and non-negated forms of garden path and unambiguous sentences. The ambiguous verb ``handed'' primes the effect later in the sentence, while the unambiguous ``given'' avoids it. Like in Figure~\ref{fig:cos_test}, all metrics catch the garden path effect at the verb ``grumbled'', but only cosine similarity and Manhattan distance anticipate the possible effect at ``littered''}
    \label{fig:trajectory2}
\end{figure}

On the other hand, the verb ``tried'' is not ambiguous in this way. The clause it might preempt, such as ``tried for murder'', would be improperly placed and awkward. Thus, the model's hidden states prior to the verb ``tried'' in the garden path and negated sentences are nearly orthogonal to each other, whereas they bear more similarity right before the verb ``lit''. Although surprisal spikes at the verb ``tried'' as well, the surprisal trajectory does not reflect the possible effect at ``lit'', illustrating the inadequacy of using this metric alone. This ambiguity, however, is only reflected in Manhattan distance and cosine similarity, and demonstrates how internal metrics can help us understand relationships between the model's syntactic states that surprisal analysis alone would miss. We argue that surprisal misses this effect because it depends entirely on the word that triggers a garden path effect rather than the state of the decoder prior to the trigger, which more holistically encodes the ambiguity that creates the environment where a garden path effect can occur. In this case, it seems that the prior likelihood of various constructions after `lit' leads GPT-2 to suspect it leads a reduced relative clause (i.e. `that was lit') and is not the sentence's main verb, but our distance metrics between the model's hidden states prior to this potential trigger show that GTP-2 nonetheless registers the possibility of a garden path effect. However, more work is needed to explore how exactly each of these metrics measures this abstract concept, and how sensitive they are to other syntactic and semantic effects.

We highlight two other examples of this phenomenon. In Figure~\ref{fig:trajectory1}, the preceding clause ``The horse raced past the barn filled [...]'' could be completed either with the actual continuation ``with tractors [...]'', or with a direct object for ``filled'', for instance: ``The horse raced past the barn filled her trainer with admiration''. In Figure~\ref{fig:trajectory2}, the clause ``The lawyer handed the report littered [...]'' could likewise either continue as shown, ``with profanity [...]'', or could instead be the main verb of a shorter sentence: ``The lawyer handed the report littered as he walked across the street''.
Whereas surprisal overlooks the temporary ambiguities in these examples because it relies on the next word to trigger the garden path effect, hidden state metrics reveal them because they can directly measure the ambiguity itself.

Our analysis revealed a few unexpected results. Most prominent among these is the extent to which the addition of the negating token (``that'') to unambiguous NP/S sentences leaves the hidden representation of the sentence unchanged. Across all metrics, the negated and garden path forms of NP/S sentences are closest together, showing that except in cases where it resolves a clear ambiguity, the negating token in these sentences contributes very little to the model's internal representation. The blocked form of NP/Z sentences shows a similar indifference to the negating token (in this case, the addition of a comma), which curiously does not extend to the unambiguous form.

\section{Conclusion}
This paper presents a suite of methods to analyze the internal representations of transformer decoder language models such as GPT-2, taking advantage of a richer reflection of the model's internal process than can be ascertained from the output of the language modeling head alone. We use Manhattan distance and cosine similarity between the hidden states of GPT-2 to show that the model is affected by garden path effects in ways that are predictable based on human readers' difficulty with these sentences. Although conventional surprisal analysis mirrors these effects in many cases, it exhibits higher variance than Manhattan distances, and misses certain nuances. On this basis, we argue that linguists should look to a decoder model's hidden state for a more complete picture of syntactic state than surprisal alone can report. 

Our belief is that Manhattan distance should be the preferred metric for this type of analysis, but we invite researchers to introduce and explore new metrics to challenge this hypothesis. We hope that these early insights will help inspire deeper exploration of the hidden states of decoder-only language models. Possible directions for future work could more closely examine how information is transformed across different decoder layers within GPT-2, and might explore causes for differences between Manhattan distance and cosine similarity trajectories. The methods introduced in this study can also be used to explore decoder models' handling of arbitrary syntactic phenomena beyond garden path effects, such as verb subordination or negative polarity item licensing.

\newpage

\section{Limitations}
One weakness of this type of analysis is the necessity of having side-by-side examples, with a single perturbation between them, to compare between. The beam search approach used by \citet{Aina2021TheLM} avoids this requirement, but still relies on the language modeling head, so more work is needed to integrate these benefits.
Another difficulty is the size of the dataset; although larger than all previous datasets of garden path sentences, it only includes 83 distinct sentences, and while many more variations can be generated with the scripts we include, there is substantial overlap between these that makes training a model on these challenging. Finally, since weights for GPT-3 were not available at the time we conducted this research, our analysis is constrained to the smaller GPT-2. 

\bibliography{anthology,custom}

\begin{thebibliography}{25}
\expandafter\ifx\csname natexlab\endcsname\relax\def\natexlab#1{#1}\fi

\bibitem[{Aggarwal et~al.(2001)Aggarwal, Hinneburg, and
  Keim}]{euclidean-manhattan}
Charu~C. Aggarwal, Alexander Hinneburg, and Daniel~A. Keim. 2001.
\newblock On the surprising behavior of distance metrics in high dimensional
  space.
\newblock In \emph{Database Theory --- ICDT 2001}, pages 420--434, Berlin,
  Heidelberg. Springer Berlin Heidelberg.

\bibitem[{Aina and Linzen(2021)}]{Aina2021TheLM}
Laura Aina and Tal Linzen. 2021.
\newblock \href {http://arxiv.org/abs/2109.07848} {The language model
  understood the prompt was ambiguous: Probing syntactic uncertainty through
  generation}.
\newblock \emph{CoRR}, abs/2109.07848.

\bibitem[{Brown et~al.(2020)Brown, Mann, Ryder, Subbiah, Kaplan, Dhariwal,
  Neelakantan, Shyam, Sastry, Askell, Agarwal, Herbert-Voss, Krueger, Henighan,
  Child, Ramesh, Ziegler, Wu, Winxter, Hesse, Chen, Sigler, Litwin, Gray,
  Chess, Clark, Berner, McCandlish, Radford, Sutskever, and Amodei}]{GPT3}
Tom~B. Brown, Benjamin Mann, Nick Ryder, Melanie Subbiah, Jared Kaplan,
  Prafulla Dhariwal, Arvind Neelakantan, Pranav Shyam, Girish Sastry, Amanda
  Askell, Sandhini Agarwal, Ariel Herbert-Voss, Gretchen Krueger, Tom Henighan,
  Rewon Child, Aditya Ramesh, Daniel~M. Ziegler, Jeffrey Wu, Clemens Winxter,
  Christopher Hesse, Mark Chen, Eric Sigler, Mateusz Litwin, Scott Gray,
  Benjamin Chess, Jack Clark, Christopher Berner, Sam McCandlish, Alec Radford,
  Ilya Sutskever, and Dario Amodei. 2020.
\newblock \href {https://doi.org/10.48550/ARXIV.2005.14165} {Language models
  are few-shot learners}.

\bibitem[{Cai et~al.(2021)Cai, Huang, Bian, and Church}]{gpt-isotropy}
Xingyu Cai, Jiaji Huang, Yuchen Bian, and Kenneth Church. 2021.
\newblock \href {https://openreview.net/forum?id=xYGNO86OWDH} {Isotropy in the
  contextual embedding space: Clusters and manifolds}.
\newblock In \emph{International Conference on Learning Representations}.

\bibitem[{Coenen et~al.(2019)Coenen, Reif, Yuan, Kim, Pearce, Vi{\'{e}}gas, and
  Wattenberg}]{BERT-geometry}
Andy Coenen, Emily Reif, Ann Yuan, Been Kim, Adam Pearce, Fernanda~B.
  Vi{\'{e}}gas, and Martin Wattenberg. 2019.
\newblock \href {http://arxiv.org/abs/1906.02715} {Visualizing and measuring
  the geometry of {BERT}}.
\newblock \emph{CoRR}, abs/1906.02715.

\bibitem[{Ethayarajh(2019)}]{gpt-cone}
Kawin Ethayarajh. 2019.
\newblock \href {http://arxiv.org/abs/1909.00512} {How contextual are
  contextualized word representations? comparing the geometry of bert, elmo,
  and {GPT-2} embeddings}.
\newblock \emph{CoRR}, abs/1909.00512.

\bibitem[{Futrell et~al.(2019)Futrell, Wilcox, Morita, Qian, Ballesteros, and
  Levy}]{psycholinguistic-subjects}
Richard Futrell, Ethan Wilcox, Takashi Morita, Peng Qian, Miguel Ballesteros,
  and Roger Levy. 2019.
\newblock \href {https://doi.org/10.48550/ARXIV.1903.03260} {Neural language
  models as psycholinguistic subjects: Representations of syntactic state}.

\bibitem[{Grodner et~al.(2003)Grodner, Gibson, Argaman, and
  Babyonyshev}]{Grodner2003}
Daniel Grodner, Edward Gibson, Vered Argaman, and Maria Babyonyshev. 2003.
\newblock \href {https://doi.org/10.1023/A:1022496223965} {Against repair-based
  reanalysis in sentence comprehension}.
\newblock \emph{Journal of Psycholinguistic Research}, 32(2):141--166.

\bibitem[{Heidenreich and Williams(2021)}]{gpt2-flat-earth}
Hunter Heidenreich and Jake Williams. 2021.
\newblock \href {https://doi.org/10.1145/3461702.3462578} {The earth is flat
  and the sun is not a star: The susceptibility of gpt-2 to universal
  adversarial triggers}.
\newblock pages 566--573.

\bibitem[{Hu et~al.(2020)Hu, Gauthier, Qian, Wilcox, and
  Levy}]{hu-etal-2020-systematic}
Jennifer Hu, Jon Gauthier, Peng Qian, Ethan Wilcox, and Roger Levy. 2020.
\newblock \href {https://doi.org/10.18653/v1/2020.acl-main.158} {A systematic
  assessment of syntactic generalization in neural language models}.
\newblock In \emph{Proceedings of the 58th Annual Meeting of the Association
  for Computational Linguistics}, pages 1725--1744, Online. Association for
  Computational Linguistics.

\bibitem[{Kovaleva et~al.(2019)Kovaleva, Romanov, Rogers, and
  Rumshisky}]{BERT-dark-secrets}
Olga Kovaleva, Alexey Romanov, Anna Rogers, and Anna Rumshisky. 2019.
\newblock \href {http://arxiv.org/abs/1908.08593} {Revealing the dark secrets
  of {BERT}}.
\newblock \emph{CoRR}, abs/1908.08593.

\bibitem[{Liu et~al.(2019)Liu, Gardner, Belinkov, Peters, and
  Smith}]{linguistic-knowledge-contextual-representation}
Nelson~F. Liu, Matt Gardner, Yonatan Belinkov, Matthew~E. Peters, and Noah~A.
  Smith. 2019.
\newblock \href {https://doi.org/10.18653/v1/N19-1112} {Linguistic knowledge
  and transferability of contextual representations}.
\newblock In \emph{Proceedings of the 2019 Conference of the North {A}merican
  Chapter of the Association for Computational Linguistics: Human Language
  Technologies, Volume 1 (Long and Short Papers)}, pages 1073--1094,
  Minneapolis, Minnesota. Association for Computational Linguistics.

\bibitem[{Rogers et~al.(2020)Rogers, Kovaleva, and
  Rumshisky}]{BERTology-primer}
Anna Rogers, Olga Kovaleva, and Anna Rumshisky. 2020.
\newblock \href {http://arxiv.org/abs/2002.12327} {A primer in bertology: What
  we know about how {BERT} works}.
\newblock \emph{CoRR}, abs/2002.12327.

\bibitem[{Rudman et~al.(2021)Rudman, Gillman, Rayne, and Eickhoff}]{isoscore}
William Rudman, Nate Gillman, Taylor Rayne, and Carsten Eickhoff. 2021.
\newblock \href {http://arxiv.org/abs/2108.07344} {Isoscore: Measuring the
  uniformity of vector space utilization}.
\newblock \emph{CoRR}, abs/2108.07344.

\bibitem[{Sarti(2020)}]{sarti-thesis}
Gabriele Sarti. 2020.
\newblock \emph{Interpreting Neural Language Models for Linguistic Complexity
  Assessment}.
\newblock Ph.D. thesis, University of Trieste.

\bibitem[{Staub(2007)}]{Staub2007-rn}
Adrian Staub. 2007.
\newblock The parser doesn't ignore intransitivity, after all.
\newblock \emph{J. Exp. Psychol. Learn. Mem. Cogn.}, 33(3):550--569.

\bibitem[{Sturt et~al.(1999)Sturt, Pickering, and Crocker}]{reading-times}
Patrick Sturt, Martin Pickering, and Matther Crocker. 1999.
\newblock Structural change and reanalysis difficulty in language
  comprehension.
\newblock \emph{Journal of Memory and Language}.

\bibitem[{Tabor and Hutchins(2004)}]{Tabor2004-ev}
Whitney Tabor and Sean Hutchins. 2004.
\newblock Evidence for self-organized sentence processing: digging-in effects.
\newblock \emph{J. Exp. Psychol. Learn. Mem. Cogn.}, 30(2):431--450.

\bibitem[{Tenney et~al.(2019{\natexlab{a}})Tenney, Das, and
  Pavlick}]{BERT-classical-nlp}
Ian Tenney, Dipanjan Das, and Ellie Pavlick. 2019{\natexlab{a}}.
\newblock \href {http://arxiv.org/abs/1905.05950} {{BERT} rediscovers the
  classical {NLP} pipeline}.
\newblock \emph{CoRR}, abs/1905.05950.

\bibitem[{Tenney et~al.(2019{\natexlab{b}})Tenney, Xia, Chen, Wang, Poliak,
  McCoy, Kim, Durme, Bowman, Das, and Pavlick}]{what-do-you-learn-from-context}
Ian Tenney, Patrick Xia, Berlin Chen, Alex Wang, Adam Poliak, R~Thomas McCoy,
  Najoung Kim, Benjamin~Van Durme, Sam Bowman, Dipanjan Das, and Ellie Pavlick.
  2019{\natexlab{b}}.
\newblock \href {https://openreview.net/forum?id=SJzSgnRcKX} {What do you learn
  from context? probing for sentence structure in contextualized word
  representations}.
\newblock In \emph{International Conference on Learning Representations}.

\bibitem[{Timkey and van Schijndel(2021)}]{gpt-rogue-dims}
William Timkey and Marten van Schijndel. 2021.
\newblock \href {http://arxiv.org/abs/2109.04404} {All bark and no bite: Rogue
  dimensions in transformer language models obscure representational quality}.
\newblock \emph{CoRR}, abs/2109.04404.

\bibitem[{van Schijndel and Linzen(2018)}]{vanSchijndel2018ModelingGP}
Marten van Schijndel and Tal Linzen. 2018.
\newblock \href {https://cogsci.mindmodeling.org/2018/papers/0496/0496.pdf}
  {Modeling garden path effects without explicit hierarchical syntax}.
\newblock \emph{Cognitive Science}.

\bibitem[{van Schijndel and Linzen(2019)}]{Schijndel2019NeuralNS}
Marten van Schijndel and Tal Linzen. 2019.
\newblock Neural network surprisal predicts the existence but not the magnitude
  of human syntactic disambiguation difficulty.

\bibitem[{van Schijndel and Linzen(2021)}]{vanSchijndel2021SingleStagePM}
Marten van Schijndel and Tal Linzen. 2021.
\newblock Single-stage prediction models do not explain the magnitude of
  syntactic disambiguation difficulty.
\newblock \emph{Cognitive science}, 45 6:e12988.

\bibitem[{Wallace et~al.(2019)Wallace, Feng, Kandpal, Gardner, and
  Singh}]{universal-adversarial-triggers}
Eric Wallace, Shi Feng, Nikhil Kandpal, Matt Gardner, and Sameer Singh. 2019.
\newblock \href {https://doi.org/10.48550/ARXIV.1908.07125} {Universal
  adversarial triggers for attacking and analyzing nlp}.

\end{thebibliography}
\bibliographystyle{acl_natbib}

\end{document}